\documentclass[runningheads]{llncs}
\usepackage{times}
\usepackage{epsfig}
\usepackage{graphicx}
\usepackage{amsmath}
\usepackage{amssymb}
\usepackage{graphicx}
\usepackage{amsmath,amssymb}
\usepackage{color}
\usepackage{graphicx}
\usepackage{comment}
\usepackage{amsmath,amssymb}
\usepackage{color}
\usepackage{graphicx}
\usepackage{subfigure}
\usepackage{booktabs}
\usepackage{float}

\newcommand{\specialcell}[2][c]{%
  \begin{tabular}[#1]{@{}c@{}}#2\end{tabular}}
\usepackage{algorithm,algpseudocode}
\usepackage{float}
\usepackage{amssymb}
\usepackage{multirow}
\usepackage[misc,geometry]{ifsym}
\newcommand{\etal}{{\em et al. }}
\usepackage{hyperref}
\usepackage[misc,geometry]{ifsym}
\usepackage{colorspace}
\definespotcolor{myblue}{PANTONE 7716 C}{.83, 0, .40, .11}

\begin{document}

\title{FEDS - Filtered Edit Distance Surrogate}

\titlerunning{FEDS - Filtered Edit Distance Surrogate}

\author{Yash Patel (\Letter) \orcidID{0000-0001-9373-529X} \and Ji{\v{r}}{\'i} Matas \orcidID{0000-0003-0863-4844}}
\authorrunning{Y. Patel \& J. Matas}
\institute{Visual Recognition Group, Czech Technical University in Prague \email{\{patelyas,matas\}@fel.cvut.cz}} 

\maketitle

\begin{abstract}
       This paper proposes a procedure to train a scene text recognition model using a robust learned surrogate of edit distance. The proposed method borrows from self-paced learning and filters out the training examples that are hard for the surrogate. The filtering is performed by judging the quality of the approximation, using a ramp function, enabling end-to-end training. Following the literature, the experiments are conducted in a post-tuning setup, where a trained scene text recognition model is tuned using the learned surrogate of edit distance. The efficacy is demonstrated by improvements on various challenging scene text datasets such as IIIT-5K, SVT, ICDAR, SVTP, and CUTE. The proposed method provides an average improvement of $11.2 \%$ on total edit distance and an error reduction of $9.5\%$ on accuracy. 
\end{abstract}

\section{Introduction}

Supervised deep learning has benefited many tasks in computer vision, to the point that these methods are now commercially used. It involves training a neural network on a task-specific dataset annotated by humans. The training is performed with a loss function that compares the model output with the expected output. The loss function choice is driven by multiple factors such as the application-defined objective, generalization to out-of-distribution samples, and constrained by the training algorithm.

\begin{figure}[H]
    \centering
    \includegraphics[width=\textwidth]{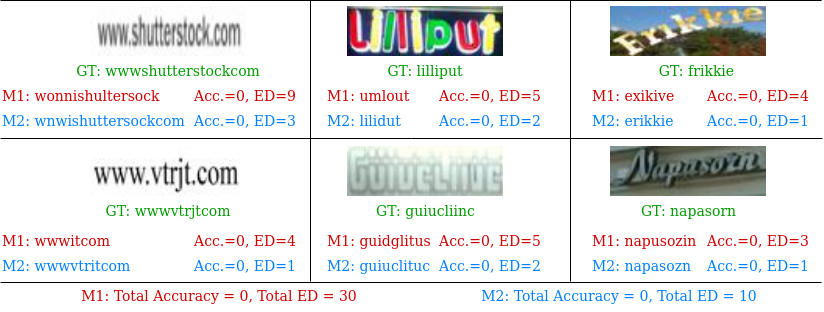}
    \caption{Accuracy and edit distance comparison for different predictions of scene text recognition (STR). For the scene text images, \textcolor{green}{green} shows the ground truth, \textcolor{red}{red} shows the prediction from a STR model \textcolor{red}{M1} and \textcolor{blue}{blue} shows the predictions from another STR model \textcolor{blue}{M2}. For these examples accuracy ranks both the models equally, however, it can be clearly seen that for the predictions in blue vocabulary search or Google search will succeed.
    }
    \label{fig:acc_vs_ed}
\end{figure}

\begin{figure}
    \centering
    \includegraphics[width=\textwidth]{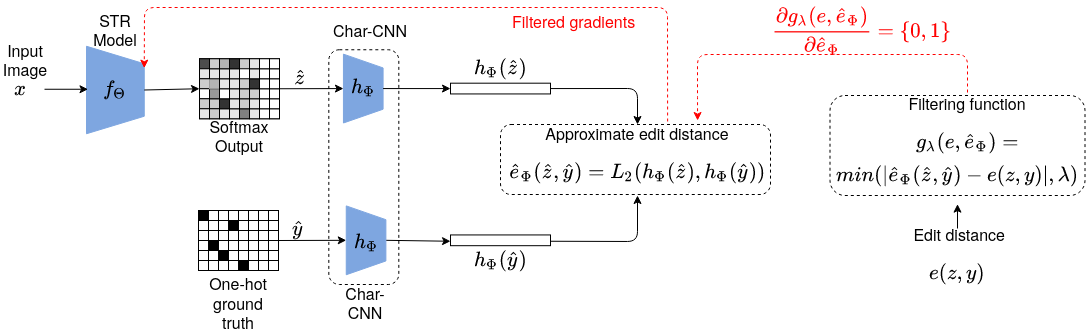}
    \caption{Overview of the proposed post-tuning procedure. $x$ is the input to the STR model $f_{\Theta}(x)$ with output $\hat{z}$. $y$ is the ground truth, $\hat{y}$ is the ground truth expressed as one-hot, $e(z,y)$ is the evaluation metric, $\hat{e}_{\Phi}(\hat{z}, \hat{y})$ is the learned surrogate and $g_{\lambda}(e, \hat{e}_{\Phi})$ is the filtering function. The approximations from the learned surrogate are checked against the edit distance by the filtering function. The STR model is not trained on the samples where the surrogate is incorrect.}
    \label{fig:feds_overall}
\end{figure}

Deep neural networks are trained by back-propagating gradients \cite{rumelhart1986learning}, which requires the loss function to be differentiable. However, the task-specific objective is often defined via an evaluation metric, which may not be differentiable. The evaluation metric's design is to fulfill the application requirements, and for the cases where the evaluation metric is differentiable, it is directly used as a loss function. For scene text recognition (STR), accuracy and edit distance are popular evaluation metric choices. Accuracy rewards the method if the prediction exactly matches the ground truth. Whereas edit distance (ED) is defined by counting addition, subtraction, and substitution operations, required to transform one string into another. As shown in Figure \ref{fig:acc_vs_ed}, accuracy does not account for partial correctness. Note that the low {\em ED} errors from M2 can be easily corrected by a dictionary search in a word-spotting setup \cite{patel2016dynamic}. Therefore, edit distance is a better metric, especially when the state-of-the-art is saturated on the benchmark datasets \cite{karatzas2013icdar,karatzas2015icdar,nayef2019icdar2019}.

When the evaluation metric is non-differentiable, a proxy loss is employed, which may not align well with the evaluation metric. Edit distance is computed via dynamic programming and is non-differentiable. Therefore, it can not be used as a loss function for training deep neural networks. The proxy loss used for training STR models is per-character cross-entropy or Connectionist Temporal Classification (CTC) \cite{graves2006connectionist}. The models trained with cross-entropy or CTC may have a sub-optimal performance on edit distance as they optimize a different objective.

The aforementioned issue can be addressed by learning a surrogate, e.g. \cite{patel2020learning}, where a model trained with the proxy loss is post-tuned on a learned surrogate of the evaluation metric. In \cite{patel2020learning}, post-tuning has shown significant improvement in performance on the evaluation metric. While Patel ~\etal \cite{patel2020learning} have paid attention to learning the surrogate, none was given to robustly train the neural network with the surrogate. In the training procedure, \cite{patel2020learning} assumes that the learned surrogate robustly estimates the edit distance for all samples. Since the surrogate is learned via supervised training, it is prone to overfitting on the training distribution and may fail on out-of-the-distribution samples. In hope for better generalizability of the surrogate, \cite{patel2020learning} makes use of a data generator to train the surrogate, which requires extra engineering effort. This paper shows that the learned edit distance surrogate often fails, leading to noisy training.

As an improvement, this paper proposes \textbf{F}iltered \textbf{E}dit \textbf{D}istance \textbf{S}urrogate. In FEDS, the STR model is trained only on the samples where the surrogate approximates the edit distance within a small error bound. This is achieved by computing the edit distance for a training sample and comparing it with the approximation from the surrogate. The comparison is realized by a ramp-function, which is piece-wise differentiable, allowing for end-to-end training. Figure \ref{fig:feds_overall} provides an overview of the proposed method. The proposed training method simplifies the training and eliminates the need for a data generator to learn a surrogate.

The rest of the paper is structured as follows. Related work is reviewed in Section~\ref{sec:related_work}, the technique for robustly training with the learned surrogate of ED is presented in Section~\ref{sec:feds}, experiments are shown in Section~\ref{sec:experiments} and the paper is concluded in Section~\ref{sec:conclusions}.

\section{Related Work}
\label{sec:related_work}

Scene text recognition (STR) is the task of recognizing text from images against complex backgrounds and layouts. STR is an active research area; comprehensive surveys can be found in \cite{ye2014text,baek2019wrong,long2020scene}. Before deep learning, STR methods focused on recognizing characters via sliding window, and hand-crafted features \cite{wang2011end,wang2010word,yao2014strokelets}. Deep learning based STR methods have made a significant stride in improving model architectures that can handle both regular (axis-aligned text) and irregular text (complex layout, such as perspective and curved text). Selected relevant methods are discussed subsequently.

\paragraph{Convolutional models for STR.} Among the first deep learning STR methods was the work of Jaderberg ~\etal \cite{jaderberg2014deep}, where a character-centric CNN \cite{lecun1998gradient} predicts a text/no-text score, a character, and a bi-gram class. Later this work was extended to word-level recognition \cite{jaderberg2016reading} where the CNN takes a fixed dimension input of the cropped word and outputs a word from a fixed dictionary. Bušta ~\etal \cite{buvsta2018e2e,busta2017deep} proposed a fully-convolutional STR model, which operates on variable-sized inputs using bi-linear sampling \cite{jaderberg2015spatial}. The model is trained jointly with a detector in a multi-task learning setup using CTC \cite{graves2006connectionist} loss. Gomez ~\etal \cite{gomez2017lsde} trains an embedding model for word-spotting, such that, the euclidean distance between the representations of two images corresponds to the edit-distance between their text strings. This embedding model differs from FEDS as it operates on images instead of STR model's predictions and is not used to train a STR model.

\paragraph{Recurrent models for STR.} Shi ~\etal \cite{shi2016end} and He ~\etal \cite{he2016reading} were among the first to propose end-to-end trainable, sequence-to-sequence models \cite{sutskeverSqeuence} for STR. An image of a cropped word is seen as a sequence of varying length, where convolutional layers are used to extract features and recurrent layers to predict a label distribution. Shi ~\etal \cite{shi2016robust} later combined the CNN-RNN hybrid with spatial transformer network \cite{jaderberg2015spatial} for better generalizability on irregular text. In \cite{shi2018aster}, Shi ~\etal adapted Thin-Plate-Spline \cite{bookstein1989principal} for STR, leading to an improved performance on both regular and irregular text (compared to \cite{shi2016robust}). While \cite{shi2018aster,shi2016robust} rectify the entire text image, Liu ~\etal \cite{liu2018char} detects and rectifies each character. This is achieved via a recurrent RoIWarp layer, which sequentially attends to a region of the feature map that corresponds to a character. Li ~\etal \cite{li2019show} passed the visual features through an attention module before decoding via an LSTM. MaskTextSpotter \cite{liao2019mask} solves detection and recognition jointly; the STR module consists of two branches while the first uses local visual features, the second utilizes contextual information in the form of attention. Litman ~\etal \cite{litman2020scatter} utilizes a stacked block architecture with intermediate supervision during training, which improves the encoding of contextual dependencies, thus improving the performance on the irregular text.

\paragraph{Training data.} Annotating scene text data in real images is complex and expensive. As an alternative, STR methods often use synthetically generated data for training. Jaderberg ~\etal \cite{jaderberg2014deep} generated $8.9$ million images by rendering fonts, coloring the image layers, applying random perspective distortion, and blending it to a background. Gupta ~\etal \cite{gupta2016synthetic} placed rendered text on natural scene images; this is achieved by identifying plausible text regions using depth and segmentation information. Patel ~\etal \cite{patel12018e2e} further extended this to multi-lingual text. The dataset of \cite{gupta2016synthetic} was proposed for training scene text detection; however, it is also useful for improving STR models \cite{baek2019wrong}. Long ~\etal \cite{long2020unrealtext} used a 3D graphics engine to generate scene text data. The 3D synthetic engine allows for better text region proposals as scene information such as normal and objects meshes are available. Their analysis shows that compared to \cite{gupta2016synthetic}, more realistic looking diverse images (contains shadow, illumination variations, etc.) are more useful for STR models. As an alternative to synthetically generate data, Janouskova ~\etal \cite{klara} leverages weakly annotated images to generate pseudo scene text labels. The approach uses an end-to-end scene text model to generate initial labels, followed by a heuristic neighborhood search to match imprecise transcriptions with weak annotations.

As discussed, significant work has been done towards improving the model architectures \cite{litman2020scatter,baek2019wrong,shi2016end,shi2016robust,shi2018aster,jaderberg2015spatial,jaderberg2016reading,busta2017deep,buvsta2018e2e,liao2019scene,zhan2019esir,yang2019symmetry,wang2020decoupled,yu2020towards,qiao2020seed,yue2020robustscanner} and obtaining data for training \cite{klara,jaderberg2016reading,gupta2016synthetic,long2020unrealtext,gomez2019selective}.

Limited attention has been paid to the loss function. Most deep learning based STR methods rely on per-character cross-entropy or CTC loss functions \cite{graves2006connectionist,baek2019wrong}. While in theory and under an assumption of infinite training data, these loss functions align with accuracy \cite{lapin2016loss}, there is no concrete evidence of their alignment with edit-distance. In comparison to the related work, this paper makes an orthogonal contribution, building upon learning surrogates \cite{patel2020learning}, this paper proposes a robust training procedure for better optimization of STR models on edit distance.

\section{FEDS: Filtered Edit Distance Surrogate }
\label{sec:feds}

\subsection{Background}
\label{sec:backgound}

The samples for training the scene text recognition (STR) model are drawn from a distribution $(x,y) \sim U_{D}$. Here, $x$ is the image of a cropped word, and $y$ is the corresponding transcription. An end-to-end trainable deep model for STR, denoted by $f_{\Theta}(x)$ predicts a soft-max output $\hat{z}=f_{\Theta}(x)$, $f_{\Theta}:\mathbb{R}^{W\times H \times 1}\xrightarrow{}\mathbb{R}^{|A|\times L}$. Here $W$ and $H$ are the dimensions of the input image, $A$ is the set of characters, and $L$ is the maximum possible length of the word.

For training, the ground truth $y$ is converted to one-hot representation $\hat{y}^{|A| \times L}$. Cross entropy (CE) is a popular choice of the loss function \cite{baek2019wrong}, which provides the loss for each character:
\begin{equation}
    \text{{\em CE}}(\hat{z}, \hat{y}) = - \frac{1}{L|A|} \sum_{i=1}^{L} \sum_{j=1}^{|A|}\hat{y}_{i,j} log(\hat{z}_{i,j})
\end{equation}

Patel ~\etal \cite{patel2020learning} learns the surrogate of edit distance via a learned deep embedding $h_{\Phi}$, where the Euclidean distance between the prediction and the ground truth corresponds to the value of the edit distance, which provides the edit distance surrogate, denoted by $\hat{e}_{\Phi}$:
\begin{equation}
    \hat{e}_{\Phi}(\hat{z}, \hat{y}) = \left\Vert h_{\Phi}(\hat{z}) - h_{\Phi}(\hat{y}) \right\Vert_{2}
    \label{eq:defination_ehat}
\end{equation}

where $h_{\Phi}$ is the Char-CNN \cite{zhang2015character,patel2020learning} with parameters $\Phi$. Note that the edit distance surrogate is defined on the one-hot representation of the ground truth and the soft-max prediction from the STR model.

\subsection{Learning edit distance surrogate}
\label{sec:lsed}

\subsubsection{Objective.}
To fairly demonstrate the improvements using the proposed FEDS, the loss for learning the surrogate is the same as LS-ED \cite{patel2020learning}:
\begin{enumerate}
    \item The learned edit distance surrogate should correspond to the value of the edit distance: \begin{equation}
        \hat{e}_{\Phi}(\hat{z}, \hat{y}) \approx e(z,y)
    \end{equation}
    where $e(z,y)$ is the edit distance defined on the string representation of the prediction and the ground truth.
    \item The first order derivative of the learned edit distance surrogate with respect to the STR model prediction $\hat{z}$ is close to $1$: \begin{equation}
       \left\Vert \frac{\partial \hat{e}_{\Phi}(\hat{z},\hat{y})}{\partial \hat{z}} \right\Vert_{2} \approx 1
       \label{eq:gp_term}
    \end{equation}
\end{enumerate}

Bounding the gradients (Equation~\ref{eq:gp_term}) has shown to enhance the training stability for Generative Adversarial Networks~\cite{gulrajani2017improved} and has shown to be useful for learning the surrogate \cite{patel2020learning}.

Both objectives are realized and linearly combined in the training loss:
\begin{equation}
    \text{loss}(\hat{z},\hat{y}) = w_{1}\left\Vert \big(\hat{e}_{\Phi}(\hat{z},\hat{y}) - e(z,y)\right\Vert_2^2 + w_{2} \left(\left\Vert \frac{\partial \hat{e}_{\Phi}(\hat{z},\hat{y})}{\partial \hat{z}} \right\Vert_{2} - 1\right)^{2}
    \label{eq:lsl_loss}
\end{equation}

\subsubsection{Training data.}
Patel ~\etal \cite{patel2020learning} uses two sources of data for learning the surrogate - the pre-trained STR model and a random generator. The random generator provides a pair of words and their edit distance and ensures uniform sampling in the range of the edit distance. The random generator helps the surrogate to generalize better, leading to an improvement in the final performance of the STR model.

The proposed FEDS does not make use of a random generator, reducing the effort and the computational cost. FEDS learns the edit distance surrogate only on the samples obtained from the STR model:
\begin{equation}
    (\hat{z},\hat{y}) \sim f_{\Theta}(x) \; | \; (x, y) \sim U_{D}
\end{equation}

\subsection{Robust Training}
\label{sec:robust_training}

The filtering function $g_{\lambda}$ is defined on the surrogate and the edit distance, parameterized by a scalar $\lambda$ that acts as a threshold to determine the quality of the approximation from the surrogate. The filtering function is defined as:

\begin{equation}
    g_{\lambda}(e(z,y),\hat{e}_{\Phi}(\hat{z}, \hat{y})) = \min(|\hat{e}_{\Phi}(\hat{z}, \hat{y}) - e(z,y)|, \lambda) \; | \; \lambda > 0
    \label{eq:filtering_fuc}
\end{equation}

\begin{figure}
    \centering
    \includegraphics[width=0.6\textwidth]{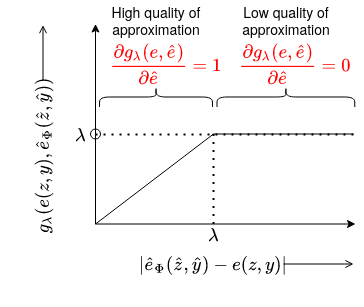}
    \caption{The filtering function enforces zero gradients for the samples that are hard for the surrogate (low quality of approximation). STR model is trained only on the samples where the quality of the approximation from the edit distance surrogate is high.}
    \label{fig:g_lambda}
\end{figure}

The filtering function is piece-wise differentiable, as can be seen in Figure \ref{fig:g_lambda}. For the samples where the quality of approximation from the surrogate is low, the gradients are zero, and the STR model is not trained on those samples. Whereas for samples where the quality of the approximation is within the bound of $\lambda$, the STR model is trained to minimize the edit distance surrogate.

Learning of the ED surrogate $\hat{e}_{\Phi}$ and post-tuning of the STR model $f_{\Theta}(x)$ are conducted alternatively. The surrogate is learned first for $I_{a}$ number of iterations while the STR model is fixed. Subsequently, the STR model is trained using the surrogate and the filtering function, while the ED surrogate parameters are kept fixed. Algorithm \ref{alg:feds} and Figure \ref{fig:feds_overall} demonstrate the overall training procedure with FEDS.

\begin{algorithm}[t!]
\caption{Post-tuning with FEDS}
\label{alg:feds}
\textbf{Inputs}: Supervised data $D$, evaluation metric $e$.\\
\textbf{Hyper-parameters}: Number of update steps $I_{a}$ and $I_{b}$, learning rates $\eta_{a}$ and $\eta_{b}$, number of epochs $E$.\\
\textbf{Objective}: Robustly post-tune the STR model, {\em i.e.}, $f_{\Theta}(x)$ and learn the edit distance surrogate, \emph{i.e.}, $\hat{e}_{\Phi}$.
\begin{algorithmic}[1]
\State \textit{Initialize} $\Theta \leftarrow$ pre-trained weights, $\Phi \leftarrow$ random weights.
\For{epoch = 1,...,E}
    \For{i = 1,...,$I_{a}$}
        \State sample, $(x,y) \sim U_{D}$
        \State inference, $\hat{z}=f_{\Theta^{epoch-1}}(x)$
        \State compute loss, $l_{\hat{e}} =$ {\em loss}$(\hat{z}, \hat{y})$ (Equation \ref{eq:lsl_loss})
        \State update ED surrogate, $\Phi^{i} \leftarrow \Phi^{i-1} - \eta_{a} \frac{\partial l_{\hat{e}}}{\partial \Phi^{i-1}}$
    \EndFor \\
   \quad \quad $\Phi \leftarrow \Phi^{I_{a}}$
    \For{i = 1,...,$I_{b}$}
        \State sample, $(x,y) \sim U_{D}$
        \State inference, $\hat{z} = f_{\Theta^{i-1}}(x)$
        \State compute ED from the surrogate, $\hat{e} = \hat{e}_{\Phi^{epoch}}(\hat{z}, \hat{y})$ (Equation \ref{eq:defination_ehat})
        \State compute ED, $e = e(z, y)$
        \State computer loss, $l_{f} = g_{\lambda}(e, \hat{e})$ (Equation \ref{eq:filtering_fuc})
        \State update STR model, $\Theta^{i} \leftarrow \Theta^{i-1} - \eta_{b} \frac{\partial(l_{f})}{\partial \Theta^{i-1}}$
    \EndFor \\
    \quad \quad $\Theta \leftarrow \Theta^{I_{b}}$
\EndFor
\end{algorithmic}
\end{algorithm}

\section{Experiments}
\label{sec:experiments}

\subsection{FEDS model}
\label{sec:feds_architecture}

The model for learning the deep embedding, {\em i.e.}, $h_{\Phi}$ is kept same as \cite{patel2020learning}. A Char-CNN architecture~\cite{zhang2015character} is used with five $1D$ convolution layers, LeakyReLU \cite{DBLP:booktitles/corr/XuWCL15} and two {\em FC} layers. The embedding model, $h_{\Phi}$, maps the input to a $1024$ dimensions, $h_{\Phi}: \mathbb{R}^{|A|\times L} \xrightarrow{} \mathbb{R}^{1024}$. Feed forward (Equation \ref{eq:defination_ehat}), generates embeddings for the ground-truth $\hat{y}$ (one-hot) and model prediction $\hat{z}$ (soft-max) and an approximation of edit distance is computed by $L_{2}$ distance between the two embedding.

\subsection{Scene Text Recognition model}
Following the survey on STR, \cite{baek2019wrong}, the state-of-the-art model ASTER is used \cite{shi2018aster}, which contains four modules: (a) transformation, (b) feature extraction, (c) sequence modeling, and (d) prediction. Baek~\etal~\cite{baek2019wrong} provides a detailed analysis of STR models and the impact of different modules on the performance.

\subsubsection{Transformation.} Operates on the input image and rectifies the curved or tilted text, easing the recognition for the subsequent modules. The two popular variants include Spatial Transformer \cite{jaderberg2015spatial} and Thin Plain Spline (TPS) \cite{shi2018aster}. TPS employs a smooth spline interpolation between a set of fiducial points, which are fixed in number. Following the analysis of Shi ~\etal \cite{shi2018aster} \cite{baek2019wrong}, the STR model used employs TPS.

\subsubsection{Feature extraction.} Involves a Convolutional Neural Network \cite{lecun1998gradient}, that extracts the features from the image transformed by TPS. Popular choices include VGG-16 \cite{simonyan2014very} and ResNet \cite{he2016deep}. Follwoing \cite{baek2019wrong}, the STR model used employs ResNet for the ease of optimization and good performance. 

\subsubsection{Sequence modeling.} Captures the contextual information within a sequence of characters; this module operates on the features extracted from a ResNet. The STR model used employs BiLSTM \cite{hochreiter1997long}.

\subsubsection{Prediction.} The predictions are made based on the identified features of the image. The prediction module depends on the loss function used for training. CTC loss requires the prediction to by sigmoid, whereas cross-entropy requires the prediction to be a soft-max distribution over the set of characters. The design of FEDS architecture (Section \ref{sec:feds_architecture}) requires a soft-max distribution.

FEDS and LS-ED \cite{patel2020learning} are investigated with the state-of-the-art performing configuration of the STR model, which is \textit{TPS-ResNet-BiLSTM-Attn}.

\subsection{Training and Testing data}

The STR models are trained on synthetic and pseudo labeled data and are evaluated on real-world benchmarks. Note that the STR models are not fine-tuned on evaluation datasets (same as \cite{baek2019wrong}).
\subsubsection{Training data.} The experiments make use of the following synthetic and pseudo labeled data for training:
\begin{itemize}
    \item \textbf{MJSynth} \cite{jaderberg2014deep} (synthetic). $8.9$ million synthetically generated images, obtained by rendering fonts, coloring the image layers, applying random perspective distortion, and blending it to a background.
    \item \textbf{SynthText} \cite{gupta2016synthetic} (synthetic). $5.5$ million text instance by placing rendered text on natural scene images. This is achieved by identifying plausible text regions using depth and segmentation information.
    \item \textbf{Uber-Text} \cite{klara} (pseudo labels). $138$K real images from Uber-Text \cite{zhang2017uber} with pseudo labels obtained using \cite{klara}.
    \item \textbf{Amazon book covers} \cite{klara} (pseudo labels). $1.5$ million real images from amazon book covers with pseudo labels obtained using \cite{klara}.
\end{itemize}

\subsubsection{Testing data.} The models trained purely on the synthetic and pseudo labelled datasets are tested on a collection of real datasets. This includes regular scene text - IIIT-5K \cite{mishra2012scene}, SVT \cite{wang2011end}, ICDAR'03 \cite{lucas2003icdar} and ICDAR'13 \cite{karatzas2013icdar}, and irregular scene text ICDAR'15 \cite{karatzas2015icdar}, SVTP \cite{quy2013recognizing}  and CUTE \cite{risnumawan2014robust}. 

\subsection{Implementation details}
\label{sec:implementation_details}

The analysis of the proposed FEDS and LS-ED \cite{patel2020learning} is conducted for two setups of training data. First, similar to \cite{baek2019wrong}, the STR models are trained on the union of the synthetic data obtained from MJSynth \cite{jaderberg2014deep}, and SynthText \cite{gupta2016synthetic} resulting in a total of $14.4$ million training examples. Second, additional pseudo labeled data \cite{klara} is used to obtain a stronger baseline.

The STR models are first trained with the proxy loss, \emph{i.e.}, cross-entropy for $300K$ iterations with a mini-batch size of $192$. The models are optimized using ADADELTA \cite{DBLP:booktitles/corr/abs-1212-5701}. Once the training is complete, these models are tuned with FEDS (Algorithm \ref{alg:feds}) on the same training set for another $20K$ iterations. For learning the edit distance surrogate the weights in the loss (Equation \ref{eq:lsl_loss}) are set as $w_{1}=1, w_{2}=0.1$. Note that the edit distance value is a non-negative integer, therefore, optimal range for $\lambda$ is $(0,0.5)$. Small value of $\lambda$ filters out substantial number of samples, slowing down the training, whereas, large values of $\lambda$ allows a noisy training. Therefore, the threshold for the filtering function (Equation \ref{eq:filtering_fuc}) is set as $\lambda=0.25$, {\em i.e.}, in the middle of the optimal range.

\subsection{Quality of the edit distance surrogate}
\label{sec:quality_of_ed}

\begin{figure}[H]
    \centering
    \includegraphics[width=\textwidth]{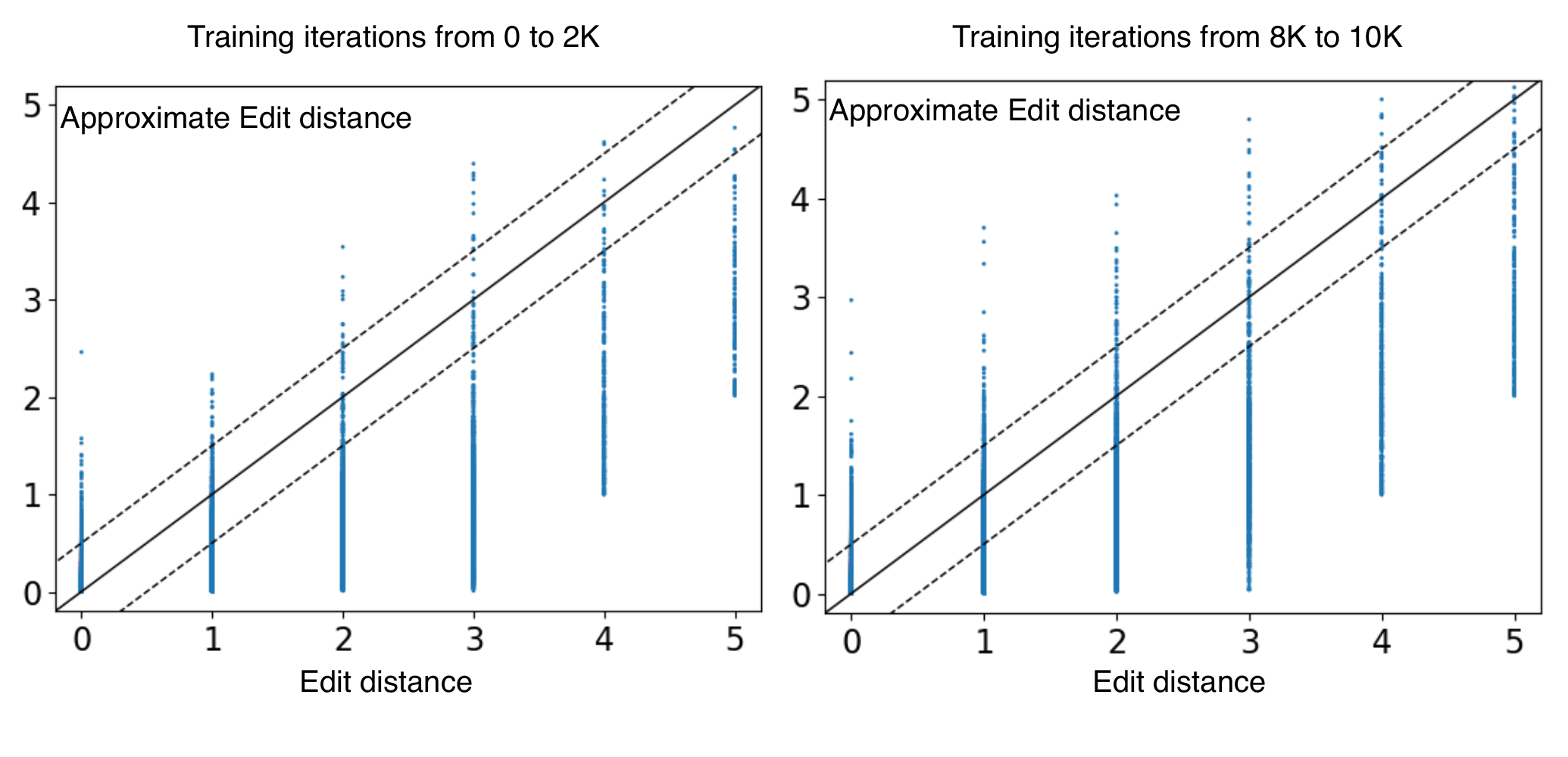}
    \caption{A comparison between the true edit distance and the approximated edit distance is shown. Each point represents a training sample for the STR model. The solid line represents an accurate approximation of the edit distance. The dotted lines represent the filtering in FEDS. \textbf{Left}: Plot for the first 2K iterations of the STR model training. \textbf{Right}: Plot for iterations from 8K to 10K of the STR model training.}
    \label{fig:e_vs_ehat}
\end{figure}

Figure \ref{fig:e_vs_ehat} shows a comparison between the edit distance and the approximation from the surrogate. As the training progresses, the approximation improves, {\em i.e.}, more samples are closer to the solid line. The dotted lines represent the filtering in FEDS, {\em i.e.}, only the samples between the dotted lines contribute to the training of the STR model. Note that the surrogate fails for a large fraction of samples; therefore, the training without the filtering (as done in LS-ED \cite{patel2020learning}) is noisy.

\subsection{Quantitative results}
\label{sec:quant_results}

Table \ref{table:tps_resnet_bilstm_attn} shows the results with LS-ED \cite{patel2020learning} and the proposed FEDS in compression with the standard baseline \cite{baek2019wrong,shi2018aster}. For the training, only the synthetic datasets \cite{jaderberg2014deep,gupta2016synthetic} are used. Both LS-ED \cite{patel2020learning} and FEDS improve the performance on all evaluation metrics. Most significant gains are observed on total edit distance as the surrogate approximates it. In comparison with LS-ED, significant gains are observed with the proposed FEDS. On average, FEDS provides an improvement of $11.2\%$ on the total edit distance and $0.98\%$ on accuracy (an equivalent of $9.5\%$ error reduction).

\begin{table}[t!]
\caption{STR model trained with MJSynth \cite{jaderberg2014deep} and SynthText \cite{gupta2016synthetic}. Evaluation on IIIT-5K \cite{mishra2012scene}, SVT \cite{wang2011end}, IC'03 \cite{lucas2003icdar}, IC'13 \cite{karatzas2013icdar}, IC'15 \cite{karatzas2015icdar}, SVTP \cite{quy2013recognizing}  and CUTE \cite{risnumawan2014robust}. The results are reported using accuracy \textbf{Acc.} (higher is better), normalized edit distance \textbf{NED} (higher is better) and total edit distance \textbf{TED} (lower is better). Relative gains are shown in \textcolor{blue}{blue} and relative declines in \textcolor{red}{red}.}
\begin{center}
\begin{tabular}{ c | l | l | l | l}
\toprule
\multicolumn{1}{c|}{\textbf{\specialcell{Test \\ Data}}} 
& \multicolumn{1}{c|}{\textbf{\specialcell{Loss \\ Function}}} 
& \multicolumn{1}{c|}{\textbf{$\uparrow$ \specialcell{Acc.}}} 
& \multicolumn{1}{c|}{\textbf{$\uparrow$ \specialcell{NED}}} 
& \multicolumn{1}{c}{\textbf{$\downarrow$ \specialcell{TED}}}\\
\midrule
\multirow{3}{*}{IIIT-5K (3000)}  
& Cross-Entropy \cite{baek2019wrong} 
& $87.1$ 
& $0.959$ 
& $772$ \\

  & LS-ED \cite{patel2020learning} 
  & $88.0$ \textcolor{blue}{$+1.03\%$} 
  & $0.962$ \textcolor{blue}{$+0.31\%$} 
  & $680$ \textcolor{blue}{$+11.9\%$} \\
  
  & FEDS  
  & $88.8$ \textcolor{blue}{$+1.95\%$} 
  & $0.966$ \textcolor{blue}{$+0.72\%$}
  & $591$ \textcolor{blue}{$+23.44\%$} \\
\midrule

\multirow{3}{*}{SVT (647)}  
& Cross-Entropy \cite{baek2019wrong} 
& $87.2$ 
& $0.953$ 
& $175$ \\

 & LS-ED \cite{patel2020learning} 
 & $87.3$ \textcolor{blue}{$+0.11\%$} 
 & $0.954$ \textcolor{blue}{$+0.10\%$} 
 & $161$ \textcolor{blue}{$+8.00\%$} \\

 & FEDS  
 & $88.7$ \textcolor{blue}{$+1.72\%$} 
 & $0.957$ \textcolor{blue}{$+0.41\%$} 
 & $147$ \textcolor{blue}{$+16.0\%$} \\
 
\midrule
\multirow{3}{*}{IC'03 (860)} 
& Cross-Entropy \cite{baek2019wrong} 
& $95.1$ 
& $0.981$ 
& $105$  \\

 & LS-ED \cite{patel2020learning} 
 & $95.3$ \textcolor{blue}{$+0.21\%$} 
 & $0.982$ \textcolor{blue}{$+0.10\%$} 
 & $89$ \textcolor{blue}{$+15.2\%$} \\

 & FEDS  
 & $95.4$ \textcolor{blue}{$+0.31\%$} 
 & $0.983$ \textcolor{blue}{$+0.20\%$} 
 & $87$ \textcolor{blue}{$+17.1\%$} \\
 
\midrule
\multirow{3}{*}{IC'03 (867)} 
& Cross-Entropy \cite{baek2019wrong} 
& $95.1$ 
& $0.982$ 
& $102$ \\

 & LS-ED  \cite{patel2020learning} 
 & $95.2$ \textcolor{blue}{$+0.10\%$} 
 & $0.983$ \textcolor{blue}{$+0.10\%$} 
 & $90$ \textcolor{blue}{$+11.7\%$} \\

 & FEDS  
 & $95.0$ \textcolor{red}{$-0.10\%$} 
 & $0.981$ \textcolor{red}{$-0.10\%$} 
 & $81$ \textcolor{blue}{$+20.5\%$} \\
\midrule

\multirow{3}{*}{IC'13 (857)} 
& Cross-Entropy \cite{baek2019wrong} 
& $92.9$ 
& $0.979$ 
& $110$ \\

 & LS-ED \cite{patel2020learning} 
 & $93.9$ \textcolor{blue}{$+1.07\%$} 
 & $0.981$ \textcolor{blue}{$+0.20\%$} 
 & $97$ \textcolor{blue}{$+11.8\%$} \\

 & FEDS 
 & $93.8$ \textcolor{blue}{$+0.96\%$} 
 & $0.985$ \textcolor{blue}{$+0.61\%$} 
 & $99$ \textcolor{blue}{$+10.0\%$} \\

\midrule
\multirow{3}{*}{IC'13 (1015)} 
& Cross-Entropy \cite{baek2019wrong} 
& $92.2$ 
& $0.966$ 
& $140$ \\

 & LS-ED \cite{patel2020learning} 
 & $93.1$ \textcolor{blue}{$+0.97\%$} 
 & $0.969$ \textcolor{blue}{$+0.31\%$} 
 & $123$ \textcolor{blue}{$+12.1\%$} \\
 
 & FEDS 
 & $92.6$ \textcolor{blue}{$+0.43\%$} 
 & $0.969$ \textcolor{blue}{$+0.31\%$} 
 & $118$ \textcolor{blue}{$+15.7\%$} \\

\midrule
\multirow{3}{*}{IC'15 (1811)} 
& Cross-Entropy \cite{baek2019wrong} 
& $77.9$ 
& $0.915$ 
& $880$ \\

 & LS-ED \cite{patel2020learning} 
 & $78.2$ \textcolor{blue}{$+0.38\%$} 
 & $0.915$ - 
 & $851$ \textcolor{blue}{$+3.29\%$}\\

 & FEDS  
 & $78.5$ \textcolor{blue}{$+0.77\%$} 
 & $0.919$ \textcolor{blue}{$+0.43\%$} 
 & $820$ \textcolor{blue}{$+6.81\%$}\\

\midrule
\multirow{3}{*}{IC'15 (2077)} 
& Cross-Entropy \cite{baek2019wrong} 
& $75.0$ 
& $0.884$ 
& $1234$ \\

 & LS-ED \cite{patel2020learning} 
 & $75.3$ \textcolor{blue}{$+0.39\%$} 
 & $0.883$ \textcolor{red}{$-0.11\%$} 
 & $1210$ \textcolor{blue}{$+1.94\%$} \\
 
 & FEDS  
 & $75.7$ \textcolor{blue}{$+0.93\%$} 
 & $0.888$ \textcolor{blue}{$+0.45\%$} 
 & $1176$ \textcolor{blue}{$+4.70\%$}\\

\midrule
\multirow{3}{*}{SVTP (645)} 
& Cross-Entropy \cite{baek2019wrong} 
& $79.2$
& $0.912$ 
& $340$\\

 & LS-ED  \cite{patel2020learning} 
 & $80.0$ \textcolor{blue}{$+1.01\%$} 
 & $0.915$ \textcolor{blue}{$+0.32\%$} 
 & $327$ \textcolor{blue}{$+3.82\%$}\\

 & FEDS  
 & $80.9$ \textcolor{blue}{$+2.14\%$} 
 & $0.919$ \textcolor{blue}{$+0.76\%$} 
 & $307$ \textcolor{blue}{+$9.70\%$}\\

\midrule
\multirow{3}{*}{CUTE (288)} 
& Cross-Entropy \cite{baek2019wrong} 
& $74.9$
& $0.881$ 
& $221$\\

 & LS-ED \cite{patel2020learning} 
 & $75.6$ \textcolor{blue}{$+0.93\%$} 
 & $0.885$ \textcolor{blue}{$+0.45\%$} 
 & $204$ \textcolor{blue}{$+7.69\%$}\\ 
 
 & FEDS 
 & $75.3$ \textcolor{blue}{$+0.53\%$} 
 & $0.891$ \textcolor{blue}{$+1.13\%$} 
 & $197$ \textcolor{blue}{$+10.8\%$}\\

\midrule
\multirow{3}{*}{TOTAL} 
& Cross-Entropy \cite{baek2019wrong} 
& $85.6$
& $0.941$
& $4079$\\  

 & LS-ED \cite{patel2020learning}
 & $86.1$ \textcolor{blue}{$+0.61\%$} 
 & $0.942$ \textcolor{blue}{$+0.18\%$}
 & $3832$ \textcolor{blue}{$+6.05\%$}\\
 
 & FEDS 
 & $86.5$ \textcolor{blue}{$+0.98\%$}
 & $0.946$ \textcolor{blue}{$+0.48\%$} 
 & $3623$ \textcolor{blue}{$+11.2\%$}\\

\bottomrule
\end{tabular}
\end{center}
\label{table:tps_resnet_bilstm_attn}
\end{table}

Table \ref{table:tps_resnet_bilstm_attn_pgt} presents the results with LS-ED \cite{patel2020learning} and FEDS in compression with a stronger baseline \cite{klara}. For the training, a combination of synthetic \cite{jaderberg2014deep,gupta2016synthetic} and pseudo labelled \cite{klara} data is used. LS-ED \cite{patel2020learning} provides a limited improvement of $2.91\%$ on total edit distance whereas FEDS provides a significant improvement of $7.90\%$ and an improvement of $1.01\%$ on accuracy (equivalently $7.9\%$ error reduction). Furthermore, LS-ED \cite{patel2020learning} declines the performance on ICDAR'03 \cite{lucas2003icdar} dataset.

\begin{table}[t!]
\caption{STR model trained with MJSynth \cite{jaderberg2014deep}, SynthText \cite{gupta2016synthetic} and pseudo labelled \cite{klara} data. Evaluation on IIIT-5K \cite{mishra2012scene}, SVT \cite{wang2011end}, IC'03 \cite{lucas2003icdar}, IC'13 \cite{karatzas2013icdar}, IC'15 \cite{karatzas2015icdar}, SVTP \cite{quy2013recognizing}  and CUTE \cite{risnumawan2014robust}. The results are reported using accuracy \textbf{Acc.} (higher is better), normalized edit distance \textbf{NED} (higher is better) and total edit distance \textbf{TED} (lower is better). Relative gains are shown in \textcolor{blue}{blue} and relative declines in \textcolor{red}{red}.}
\begin{center}
\begin{tabular}{ c | l | l | l | l}
\toprule
\multicolumn{1}{c|}{\textbf{\specialcell{Test \\ Data}}} 
& \multicolumn{1}{c|}{\textbf{\specialcell{Loss \\ Function}}} 
& \multicolumn{1}{c|}{\textbf{$\uparrow$ \specialcell{Acc.}}} 
& \multicolumn{1}{c|}{\textbf{$\uparrow$ \specialcell{NED}}} 
& \multicolumn{1}{c}{\textbf{$\downarrow$ \specialcell{TED}}}\\
\midrule
\multirow{3}{*}{IIIT-5K (3000)}  
& Cross-Entropy \cite{baek2019wrong} 
& $91.7$ 
& $0.973$ 
& $550$\\
  
  & LS-ED \cite{patel2020learning} 
  & $91.8$ \textcolor{blue}{$+0.14\%$} 
  & $0.973$ - 
  & $539$ \textcolor{blue}{$+2.00\%$}\\
  
  & FEDS  
  & $92.2$ \textcolor{blue}{$+0.54\%$} 
  & $0.975$ \textcolor{blue}{$+0.20\%$} 
  & $479$ \textcolor{blue}{$+12.9\%$}\\
\midrule

\multirow{3}{*}{SVT (647)}  
& Cross-Entropy \cite{baek2019wrong} 
& $91.8$ 
& $0.970$ 
& $107$\\

 & LS-ED \cite{patel2020learning} 
 & $91.8$ - 
 & $0.971$ \textcolor{blue}{$+0.10\%$} 
 & $100$ \textcolor{blue}{$+6.54\%$}\\
 
 & FEDS  
 & $92.1$ \textcolor{blue}{$+0.32\%$}
 & $0.971$ \textcolor{blue}{$+0.10\%$} 
 & $102$ \textcolor{blue}{$+4.67\%$}\\
 
\midrule
\multirow{3}{*}{IC'03 (860)} 
& Cross-Entropy \cite{baek2019wrong} 
& $95.6$ 
& $0.984$ 
& $85$\\

 & LS-ED \cite{patel2020learning} 
 & $95.5$ \textcolor{red}{$-0.01\%$} 
 & $0.983$ \textcolor{red}{$-0.10\%$} 
 & $91$ \textcolor{red}{$-7.05\%$}\\
 
 & FEDS  
 & $96.2$ \textcolor{blue}{$+0.62\%$} 
 & $0.986$ \textcolor{blue}{$+0.20\%$} 
 & $73$ \textcolor{blue}{$+14.1\%$}\\
 
\midrule
\multirow{3}{*}{IC'03 (867)} 
& Cross-Entropy \cite{baek2019wrong} 
& $95.7$ 
& $0.984$ 
& $89$\\

 & LS-ED  \cite{patel2020learning} 
 & $95.7$ \textcolor{blue}{$+0.03\%$} 
 & $0.984$ - 
 &  $91$ \textcolor{red}{$-2.24\%$}\\
 
 & FEDS  
 & $96.4$ \textcolor{blue}{$+0.73\%$} 
 & $0.987$ \textcolor{blue}{$+0.30\%$} 
 & $77$ \textcolor{blue}{$+13.4\%$}\\
\midrule

\multirow{3}{*}{IC'13 (857)} 
& Cross-Entropy \cite{baek2019wrong} 
& $95.4$ 
& $0.988$ 
& $65$\\

 & LS-ED \cite{patel2020learning} 
 & $96.3$ \textcolor{blue}{$+1.03\%$} 
 & $0.989$ \textcolor{blue}{$+0.10\%$} 
 &  $55$ \textcolor{blue}{$+15.3\%$}\\
 
 & FEDS 
 & $96.5$ \textcolor{blue}{$+1.15\%$} 
 & $0.989$ \textcolor{blue}{$+0.10\%$} 
 &  $57$ \textcolor{blue}{$+12.3\%$}\\
 
\midrule
\multirow{3}{*}{IC'13 (1015)} 
& Cross-Entropy \cite{baek2019wrong} 
& $94.1$ 
& $0.975$ 
& $97$ \\

 & LS-ED \cite{patel2020learning} 
 & $94.8$ \textcolor{blue}{$+0.82\%$} 
 & $0.975$ - 
 &  $87$ \textcolor{blue}{$+10.3\%$}\\
 
 & FEDS 
 & $95.3$ \textcolor{blue}{$+1.27\%$} 
 & $0.975$ -
 &  $90$ \textcolor{blue}{$+7.21\%$}\\
 
\midrule
\multirow{3}{*}{IC'15 (1811)} 
& Cross-Entropy \cite{baek2019wrong} 
& $82.8$ 
& $0.939$ 
& $614$\\

 & LS-ED \cite{patel2020learning} 
 & $83.2$ \textcolor{blue}{$+0.56\%$} 
 & $0.939$ - 
 & $599$ \textcolor{blue}{$+2.44\%$}\\
 
 & FEDS  
 & $83.8$ \textcolor{blue}{$+1.20\%$} 
 & $0.942$ \textcolor{blue}{$+0.31\%$} 
 &  $578$ \textcolor{blue}{$+5.86\%$}\\
 
\midrule
\multirow{3}{*}{IC'15 (2077)} 
& Cross-Entropy \cite{baek2019wrong} 
& $80.0$ 
& $0.908$ 
& $961$\\

 & LS-ED \cite{patel2020learning} 
 & $80.4$ \textcolor{blue}{$+0.54\%$} 
 & $0.908$ - 
 &  $944$ \textcolor{blue}{$+1.76\%$} \\
 
 & FEDS  
 & $80.9$ \textcolor{blue}{$+1.12\%$} 
 & $0.91$ \textcolor{blue}{$+0.22\%$} 
 & $929$ \textcolor{blue}{$+3.32\%$}\\
 
\midrule
\multirow{3}{*}{SVTP (645)} 
& Cross-Entropy \cite{baek2019wrong} 
& $82.4$ 
& $0.930$ 
&  $271$\\

 & LS-ED  \cite{patel2020learning} 
 & $83.4$ \textcolor{blue}{$+1.22\%$} 
 & $0.933$ \textcolor{blue}{$+0.32\%$} 
 & $258$ \textcolor{blue}{$+4.79\%$}\\
 
 & FEDS  
 & $84.0$ \textcolor{blue}{$+1.94\%$} 
 & $0.935$ \textcolor{blue}{$+0.53\%$} 
 & $248$ \textcolor{blue}{$+8.48\%$}\\
 
\midrule
\multirow{3}{*}{CUTE (288)} 
& Cross-Entropy \cite{baek2019wrong} 
& $77.3$  
& $0.883$  
&  $211$\\

 & LS-ED \cite{patel2020learning} 
 & $77.3$ \textcolor{blue}{$+0.06\%$} 
 & $0.885$ \textcolor{blue}{$+0.22\%$}
 & $197$ \textcolor{blue}{$+6.63\%$}\\
 
 & FEDS 
 & $79.0$ \textcolor{blue}{$+2.19\%$} 
 & $0.898$ \textcolor{blue}{$+1.69\%$} 
 & $176$ \textcolor{blue}{$+16.5\%$}\\
 
\midrule
\multirow{3}{*}{TOTAL} 
& Cross-Entropy \cite{baek2019wrong} 
& $88.7$ 
& $0.953$
& $3050$\\

 & LS-ED \cite{patel2020learning}
 & $89.0$ \textcolor{blue}{$+0.41\%$} 
 & $0.954$ \textcolor{blue}{$+0.62\%$}
 & $2961$ \textcolor{blue}{$+2.91\%$}\\
 
 & FEDS 
 & $89.6$ \textcolor{blue}{$+1.01\%$} 
 & $0.956$ \textcolor{blue}{$+0.35\%$}
 & $2809$ \textcolor{blue}{$+7.90\%$}\\
 
\bottomrule
\end{tabular}
\end{center}
\label{table:tps_resnet_bilstm_attn_pgt}
\end{table}

\subsection{Qualitative results}
\label{sec:qual_results}

\begin{figure}[H]
    \centering
    \includegraphics[width=\textwidth]{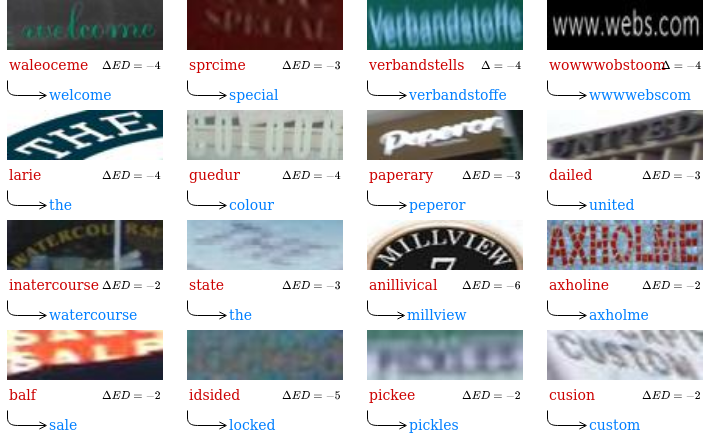}
    \caption{Randomly chosen examples from the test set where FEDS improves the STR model trained with cross-entropy. \textcolor{red}{Red} shows the incorrect predictions from the baseline model, \textcolor{blue}{blue} shows the correct prediction after post-tuning with FEDS and the arrow indicates post-tuning with FEDS.}
    \label{fig:feds_qual_good}
\end{figure}

Figure \ref{fig:feds_qual_good} shows randomly picked qualitative examples where FEDS leads to an improvement in the edit distance. Notice that the predictions from the baseline model are incorrect in all the examples. After post-tuning with FEDS, the predictions are correct, {\em i.e.}, perfectly match with the ground truth.

\begin{figure}[H]
    \centering
    \includegraphics[width=\textwidth]{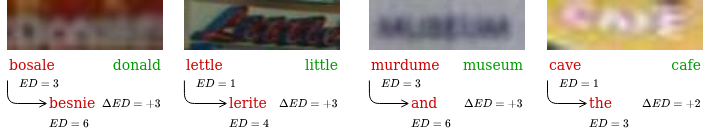}
    \caption{Four worst examples out of $12$K samples in the test set where FEDS leads to an increase in the edit distance. \textcolor{red}{Red} shows the incorrect predictions, \textcolor{green}{green} shows the ground truth and the arrow indicates post-tuning with FEDS.}
    \label{fig:feds_qual_bad}
\end{figure}

Figure \ref{fig:feds_qual_bad} shows hand-picked examples where FEDS leads to a maximum increase in the edit distance (ED increases). Notice that in these examples, the predictions from the baseline model are also incorrect. Furthermore, the input images are nearly illegible for a human.

\section{Conclusions}
\label{sec:conclusions}
This paper makes an orthogonal contribution to the trend of scene text recognition progress. It proposes a method to robustly post-tune a STR model using a learned surrogate of edit distance. The empirical results demonstrate an average improvement of $11.2 \%$ on total edit distance and an error reduction of $9.5\%$ on accuracy on a standard baseline \cite{baek2019wrong}. Improvements of $7.9\%$ on total edit distance and an error reduction of $10.3\%$ on accuracy are shown on a stronger baseline \cite{klara}. \\ Project link: \href{https://yash0307.github.io/FEDS_ICDAR2021}{yash0307.github.io/FEDS\_ICDAR2021}

\section*{Acknowledgement}
This research was supported by Research Center for Informatics (project CZ.02.1.01/\-0.0/0.0/16\_019/0000765 funded by OP VVV), CTU student grant (SGS20/171/OHK3/\-3T/13), Project StratDL in the realm of COMET K1 center Software Competence Center Hagenberg and Amazon Research Award.

\bibliographystyle{splncs04}
\bibliography{main}

\end{document}